\documentclass[runningheads]{llncs}
\usepackage[T1]{fontenc}
\usepackage{graphicx}
\usepackage{amsmath}
\usepackage{todonotes}
\usepackage{mathrsfs}
\usepackage{booktabs}
\usepackage{subcaption}
\usepackage{multicol}
\usepackage{tikz}
\usetikzlibrary{calc}
\usetikzlibrary{decorations.pathmorphing}
\newcommand{\overwave}[1]{%
  \tikz[baseline=(x.base)]{
    \node[inner sep=0] (x) {$#1$};
    \draw[decorate,decoration={snake,amplitude=1pt,segment length=6pt}] 
      ($(x.north west)+(0,2pt)$) -- ($(x.north east)+(0,2pt)$);
  }%
}
\usepackage{multirow}
\usepackage{amssymb}
\usepackage{hyperref}

\newcommand{\gmt}[1]{{\color{teal}#1}}

\newcommand{\ttrain}{\mathcal{T}_\text{train}}
\newcommand{\tval}{\mathcal{T}_\text{val}}
\newcommand{\ttest}{\mathcal{T}_\text{test}}

\newcommand{\hardware}{\mathcal{H}}

\begin{document}
\title{Link Prediction or Perdition: the Seeds of Instability in Knowledge Graph Embeddings}
\titlerunning{The Seeds of Instability in Knowledge Graph Embeddings}
%
\author{Guillaume Méroué\orcidID{0009-0004-7111-3202} \and
Fabien Gandon\orcidID{0000-0003-0543-1232} \and
Pierre Monnin\orcidID{0000-0002-2017-8426}}
\authorrunning{Méroué et al.}
%
\institute{Université Côte d’Azur, Inria, CNRS, I3S, France \\
\email{firstname.lastname@inria.fr}}
\maketitle              

\begin{abstract}
Embedding models (KGEMs) constitute the main link prediction approach to complete knowledge graphs. 
Standard evaluation protocols emphasize rank-based metrics such as MRR or Hits@$K$, but usually overlook the influence of random seeds on result stability. 
Moreover, these metrics conceal potential instabilities in individual predictions and in the organization of embedding spaces.
In this work, we conduct a systematic stability analysis of multiple KGEMs across several datasets. 
We find that high-performance models actually produce divergent predictions at the triple level  and highly variable embedding spaces. 
By isolating stochastic factors (\textit{i.e.}, initialization, triple ordering, negative sampling, dropout, hardware), we show that each independently induces instability of comparable magnitude. 
Furthermore, for a given model, hyperparameter configurations with better MRR are not guaranteed to be more stable.
Moreover, voting, albeit a known remediation mechanism, only provides a limited enhancement of stability.
These findings highlight critical limitations of current benchmarking protocols, and raise concerns about the reliability of KGEMs for knowledge graph completion.
\keywords{Random Seeds  \and Machine Learning \and Knowledge Graph Completion.}
\end{abstract}

\section{Introduction}
Knowledge Graphs (KGs)~\cite{hogan2021knowledge} serve as foundational structures for a wide range of applications in both academia and industry, including question answering~\cite{question_answering}, semantic parsing~\cite{semantic_parsing}, recommendation systems~\cite{recommender}, among others~\cite{hogan2021knowledge}.
They provide a formal and explicit network of facts in many downstream application domains (e.g., bio-chemistry, geography, pharmacology, linguistics, etc.)~\cite{chen0HJLMP0T23,hogan2021knowledge,NoyGJNPT19}.
They do so by encoding structured representations of information, typically defined within standard frameworks such as RDF~\cite{Lanthaler:14:RCA}, where knowledge is represented as triples $(h, r, t)$, indicating that a relation $r$ links a head entity $h$ to a tail entity $t$.
But, despite their widespread adoption, real-world KGs are often incomplete due to their manual or semi-automatic construction process~\cite{Paulheim17}. 
For instance, in 2015, 58\% of scientists in DBpedia lacked a link to their field of expertise~\cite{KrompassISWC2015}. 
This incompleteness has motivated various refinement tasks~\cite{hogan2021knowledge,Paulheim17}.
Among them, \textit{link prediction} (LP) aims to infer missing entities in incomplete triples of the form $(h, r, ?)$ or $(?, r, t)$. Over the years, a variety of Knowledge Graph Embedding Models (KGEMs) have been proposed for LP and have now become the preponderant methods~\cite{BiswasKCDJLLMPS23,JiPCMY22}.

\begin{figure}
    \centering
    \includegraphics[width=1.0\linewidth]{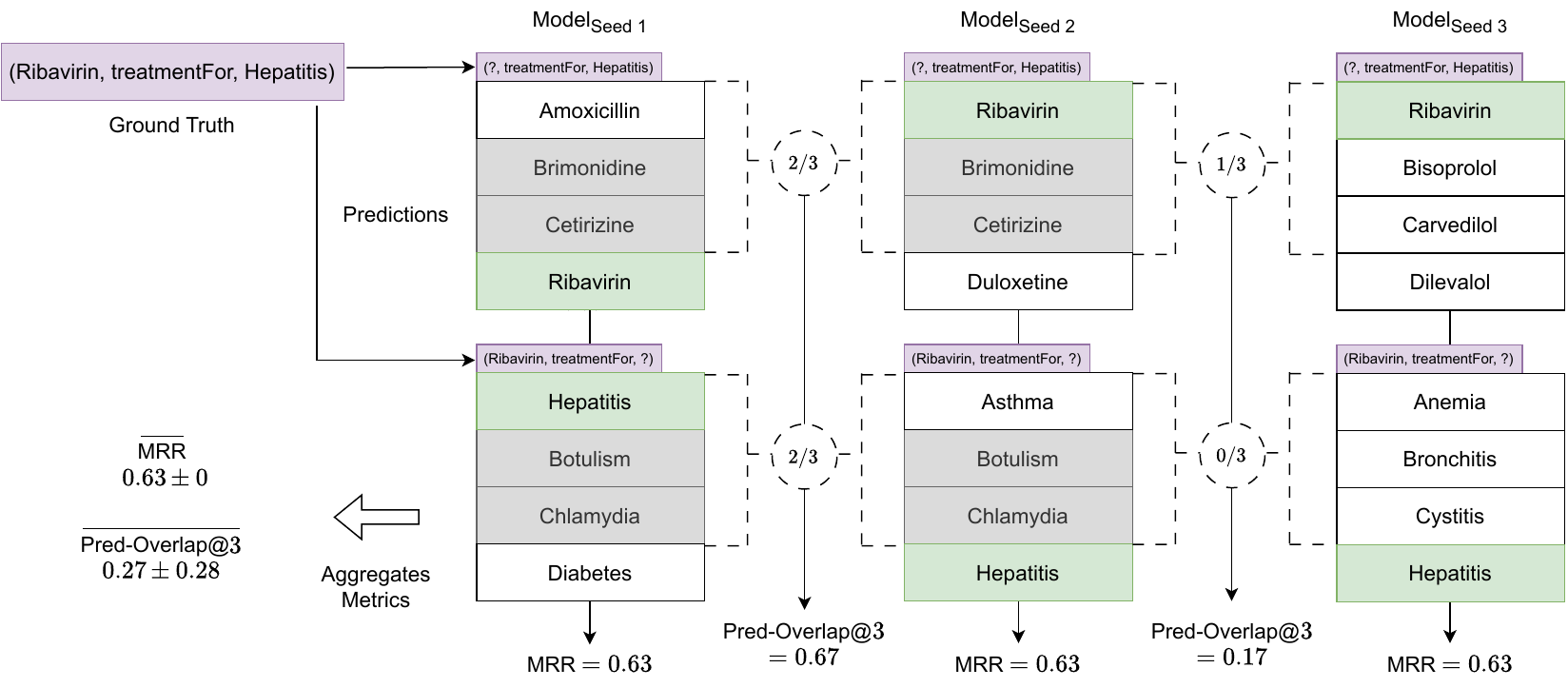}
    \caption{Illustration of three runs evaluated on two queries that yield the same MRR but with substantial differences in predictions. 
    Runs~1 and~2 rank the ground-truth entities differently but still predict largely overlapping candidate sets (Pred-Overlap@3 $= 0.67$). 
    In contrast, runs~2 and~3 assign the same rank to the ground truth, yet all other predicted entities differ, which would lead to entirely different completed graphs (Pred-Overlap@3 $= 0.17$). 
    Grey cells identify entities common to the predictions of different runs.
    }
    \label{fig:Intro}
\end{figure}

Being machine learning models, KGEMs inherently rely on stochastic elements during training, which can be controlled through random seeds. As a consequence, model performance may vary across seeds. A common good practice in machine learning is to report performance by averaging results over multiple runs with identical configurations but different seeds, and to ensure that the standard deviation remains small, indicating stable performance.
However, even when following this practice, two runs trained with different seeds may attain similar aggregate scores while nonetheless producing significantly different top-$K$ predictions.
For example, in Figure~\ref{fig:Intro}, all runs exhibit the same MRR despite displaying substantial query-by-query variability in their predictions.

Such behaviour raises a critical issue for the operational task of KG completion: under different seeds, adding the top-$K$ predicted triples to the KG may yield substantially different completed graphs. 
Even in settings where predictions are only considered as candidates rather than directly added to the graph (\textit{e.g.}, discovering drug treatments for some diseases), verifying these candidates may incur significant costs. 
In such scenarios, it is  crucial that entities are worth testing, requiring stability in predictions.
To illustrate, in Figure~\ref{fig:Intro}, completing the graph with the top-2 predictions for the query (\textit{Ribavirin, treatmentFor, ?}) using three models with identical performance may produce \textit{Hepatitis} and \textit{Bronchitis}, \textit{Asthma} and \textit{Bronchitis}, or \textit{Anemia} and \textit{Botulism}, depending solely on the seed. 
KGEM stability is thus a critical issue for operational KG completion, where seed-dependent instability may lead to unreliable and inconsistent completions, thereby undermining the trustworthiness of completed KGs as grounding structures for downstream decision-making systems. 

Prior work~\cite{zhu2024multiplicty} has already shown that global rank-based metrics commonly used for link prediction, such as MRR or Hits@$K$, fail to capture the seed-induced variability of KGEM predictions at the triple level. These authors investigated trained model agreement on predicting the gold truth entity within the top-K, which does not fully align with the operational setting of KG completion. 
Moreover, once a KGEM has been trained, its learned embedding space is often reused for downstream tasks beyond link prediction, such as entity similarity assessment~\cite{EntitySim}, conceptual clustering~\cite{KGClustering}, or logical rule mining~\cite{RuleMining}.
These downstream tasks assume the stability of the embedding space for reliability, which, to the best of our knowledge, has never been examined with respect to randomness factors.
Hence, there is a need for more comprehensive stability measures targeting the top-$K$ candidates and the embedding space.

This motivated our work, where we question and quantify the extent to which randomness affects both the predictive behavior and the internal representations of KGEMs. In particular, our research questions are as follows:

\begin{description}
\item[RQ1.] \textbf{Are KGEMs stable across different random seeds?} 
We conducted multiple runs of the same models while varying the sources of randomness. 
We introduced metrics (\textit{e.g.}, \textit{Pred-Overlap}, \textit{Space-Overlap}) to assess the similarity of predictions at the triple level, and the similarity of the learned embedding space.  
Our experimental analysis reveals that, despite apparent stability in aggregate metrics such as MRR, models exhibit pronounced variability both at the triple level and in their embedding spaces.

\item[RQ2.] \textbf{How do the different sources of randomness contribute to instability?} 

We considered the contribution of each usual source of randomness in machine learning, \textit{i.e.}, initialization, triple ordering, negative sampling, dropout, and also identified hardware (\textit{i.e.}, GPU) as an additional source of randomness. 
We showed that varying sources in isolation or jointly induces instability of comparable magnitude.

\item[RQ3.] \textbf{Is model stability correlated with link prediction performance?} 
We measured the stability of the best, median, and worst performing configurations identified in an hyperparameter search.  
While the worst-performing models are clearly highly unstable, a higher MRR does not guarantee greater stability, neither in the embedding space nor in the predicted triples.
\end{description}

We describe our proposed formalism to address these questions in Section~\ref{sec:formalism}.
The experimental protocol and results are reported in Section~\ref{sec:Results}. 
We summarise our findings and perspectives in Section~\ref{sec:conclusions}.

\section{Related Work}
\label{sec:RW}

\subsection{Knowledge Graph Embedding Models for Link Prediction}
\label{KGEM}

KGEMs encode KG entities and, in most cases, relations into a $d$-dimensional vector space with the aim of preserving as much as possible the structural regularities of the original KG while offering a compact encoding. 
In the context of link prediction, this transformation allows the model to capture implicit but statistically plausible patterns and facts, subsequently used to infer new triples. 

Link prediction (LP) with KGEMs is typically achieved by defining a scoring function $\varphi(h, r, t)$ that evaluates the plausibility of a triple $(h, r, t)$ based on the corresponding embedding vectors of $h$, $r$, and $t$. 
The inferred facts can then complete the original KG. 
Over the past decade, a wide range of KGEMs has been proposed, which can be broadly grouped into three major families.  

\textbf{Geometric methods} define the scoring function as a distance between entities after applying a geometric transformation corresponding to the relation. 
This transformation can take various forms such as a translation for TransE~\cite{TransE}, a translation in a projected space for TransR~\cite{TransR}, or a rotation for RotatE~\cite{RotatE}.  

\textbf{Tensor factorization methods} define the scoring function as a bilinear product capturing the similarity between entities under a linear transformation associated with the relation. 
The bilinear product can be restricted to a diagonal matrix as for DistMult~\cite{DistMult}, to a commuting normal matrix for ANALOGY~\cite{ANALOGY}, or extended to a complex embedding space for ComplEx~\cite{ComplEx}.  

\textbf{Neural network approaches} learn the scoring function.
Convolutional models such as ConvE~\cite{ConvE} and ConvKB~\cite{ConvKB} use convolutional layers to capture patterns in embeddings, similarly to computer vision. 
Alternatively, Graph Neural Network models such as RGCN~\cite{RGCN} or NBFNET~\cite{NBFNET} exploit message passing to aggregate information from neighboring entities. 
Transformer-based approaches either operate solely on the KG structure (like all previous models) as for Hitter~\cite{Hitter} or incorporate textual descriptions as for MoCoKGC~\cite{MoCoKGC}.

While these approaches differ in architecture and expressive power, they all share a common feature: their training process inherently involves randomness. 
In this work, we focus specifically on understanding the impact of such randomness on KGEM stability.

\subsection{Impacts of Randomness on Knowledge Graph Embedding Models}

Machine learning settings involve several factors that depend on randomness such as dataset splits, initialization of parameters, or dropout, with known impacts.
In NLP, \cite{madhyastha2019modelstabilityfunctionrandom} reports that random seeds alter attention patterns and influential words, while, in GNNs, \cite{pitfallsGNN} shows that data splits yield divergent performances.
More generally, \cite{marx2020multiplicity} defines the notion of multiplicity as the fact that similar global performance of models conceal substantial variations at the local level due to randomness factors. They also propose two metrics to quantify this phenomenon: \textit{Discrepancy} and \textit{Ambiguity}.

In the KGEM literature, the impact of randomness during training is seldom addressed. 
Section 4.3 of \cite{bonner2022understanding} studies the effect of initialization seeds and concludes that most models are robust as global metrics such as MRR exhibit low variance. 
In contrast, \cite{zhu2024multiplicty} adapts the notion of multiplicity to KGEMs and shows that models retrained with different seeds may produce different triple-level predictions, even when their overall Hits@10 scores are nearly identical.
They demonstrate that this instability can be alleviated with ensemble voting.
However, their evaluation focuses solely on whether the gold entity appears or not within the top-$K$ predictions of each model, thereby ignoring the rest of the ranked candidates.
We argue that such a ground-truth–centric perspective fails to capture the full extent of model disagreement in settings where KG completion is used operationally by adding the top-$K$ predictions to the KG. 
For instance, runs~2 and~3 in Figure~\ref{fig:Intro} would exhibit no disagreement under this evaluation, since for each query they consistently concur on whether the ground-truth entity lies inside or outside the top-3. 
Yet, despite this apparent agreement, the two models produce almost entirely different sets of predicted entities, which yield radically different completed KGs.
This highlights the need for stability measures that reveal \emph{divergence in the predicted KG completion}, beyond agreement on the position of the ground truth, aligning with the operational setting of KG completion.

Beside the stability of predictions, other works investigated the impacts of randomness on the embedding space itself.
For example, \cite{Schumacher_2021} compare the topologies of learned embedding spaces across seeds and report substantial instability, leading to unstable individual predictions even if global accuracy remain stable.
Nonetheless, their work only focuses on embedding models for homogeneous graphs.
In the KGEM literature, \cite{DoSemantic} and \cite{DoSimilar} investigate properties of embedding spaces, such as the (non-)preservation of class membership and the preservation of graph proximity. But, to the best of our knowledge, no prior work has examined the stability of these spaces with respect to randomness.

Given the growing deployment of KGEMs in operational settings, there is a pressing need for stability evaluation protocols that reflect real-world usage. 
We therefore introduce measures that quantify the similarity of KG completions produced by different KGEMs, as well as metrics that assess the consistency of their embedding spaces across random seeds.


\section{Measuring KGEM (In)stability}
\label{sec:formalism}

\subsection{Preliminaries and Notations on KGEM Training}

In our work, we consider the formal representation of a KG as a set of entities~$\mathcal{E}$, a set of relations $\mathcal{R}$, and a collection of triples $\mathcal{T} \subseteq \mathcal{E} \times \mathcal{R} \times \mathcal{E}$ encoding facts. 
Each triple $(h, r, t)$ consists of a head entity $h$, a relation $r$, and a tail entity $t$, \textit{e.g.}, $(\text{France}, \text{hasCapital}, \text{Paris})$.

We focus on KGEMs trained for LP.
In this case, $\mathcal{T}$ is split into train ($\ttrain$), validation ($\tval$), and test ($\ttest$) sets, the latter being used for model evaluation. 
Given a test triple $(h,r,t)\in\ttest$, LP consists in predicting missing entities in $(h,r,?)$ or $(?,r,t)$ by scoring and ranking all entities of the KG and expecting the gold-truth entity to be as highly ranked as possible. 
Model quality is therefore assessed over $\ttest$ with rank-based metrics such as Mean Rank (MR), Mean Reciprocal Rank (MRR), and Hits@$K$ (typically $K \in \{1,3,10\}$).
The training of KGEMs typically involves an hyperparameter tuning phase to obtain a \textbf{hyperparameter configuration} $\boldsymbol{C}$ that maximizes validation MRR or Hits@$K$.    

A less explored aspect lies in the impacts of randomness during training, associated with the choice of random seeds. 
In this work, we investigate the impact of such random factors on model results and performance. 
We identify four independent sources of randomness:
\begin{description}
    \item[Negative sampling ($\mathcal{N}$)] random corruption of positive triples to generate negative examples.
    \item[Triple ordering ($\mathcal{O}$)] random shuffling of training triples at each epoch.
    \item[Parameter initialization ($\mathcal{I}$)] random initialization of model parameters.
    \item[Dropout ($\mathcal{D}$)] random subset of neurons deactivated at each batch.
\end{description}

We define a \textbf{seed configuration} $\mathfrak{S}$ as a tuple that assigns a specific value to each source of randomness. 
For example, $
\mathfrak{S} = \{ \mathfrak{S}_\mathcal{N} = s_a,\; \mathfrak{S}_\mathcal{O} = s_b,\; \mathfrak{S}_\mathcal{I} = s_c,\; \mathfrak{S}_\mathcal{D} = s_d \}$
sets the seed $\mathfrak{S}_\mathcal{N}$ controlling negative sampling to $s_a$, and the other components follow analogously.

An \textbf{instance}
$I$ is defined as $I = \text{Instance}(\text{Dataset}, \text{KGEM}, \boldsymbol{C}, \mathfrak{S}, \hardware)$, indicating that a KGEM is trained on the specified Dataset with hyperparameter configuration $\boldsymbol{C}$, seed configuration $\mathfrak{S}$, and on hardware $\hardware$. 
In our experiments, we fix $\hardware$ to ensure identical execution conditions across all runs, and therefore omit it from the notation. 
We also omit the Dataset and KGEM when clear from context, and thus denote instances compactly by $I^{\boldsymbol{C}}_{\mathfrak{S}}$.

Let $S = \{ s_1, \dots, s_n \}$ be a set of seed values.  
We define a \textbf{comparison group} $G$ as a set of instances obtained by varying one seed of $\mathfrak{S}$ over $S$, while keeping the others fixed to $s_1$.  
For example, the group of instances where only negative sampling varies is $
G^{\boldsymbol{C}}_\mathcal{N} = \Big\{ I^{\boldsymbol{C}}_{\{\mathfrak{S}_\mathcal{N}=s_i,\; \mathfrak{S}_\mathcal{O}=s_1,\; \mathfrak{S}_\mathcal{I} = s_1,\;  \mathfrak{S}_\mathcal{D}=s_1\}} \;\Big|\; s_i \in S \Big\}$.
We define $G^{\boldsymbol{C}}_\mathcal{I}$, $G^{\boldsymbol{C}}_\mathcal{O}$, $G^{\boldsymbol{C}}_\mathcal{D}$ similarly for the other sources of randomness, and $G^{\boldsymbol{C}}_\mathcal{A}$ for the case where all seeds vary jointly over $S$.
Our objective is to determine whether predictions and embedding space organization of instances within the same group remain consistent, using spatial and prediction metrics introduced below.

\subsection{Metrics}
\label{sec:metrics}

\subsubsection{Overlap on Predictions and Embedding Space Neighbors.}

To assess the instability of a group $G^{\boldsymbol{C}}_\mathcal{X}$ for any source $\mathcal{X}$ of randomness, we rely on metrics that capture agreement either in predictions or in the embedding space.  
We adopt the top-$K$ overlap, which quantifies the intersection between two unordered sets of size $K$, ranging from $0$ (disjoint sets) to $1$ (identical sets).  
It supports straightforward interpretation, covers both the cases of automatic insertion and semi-automatic approaches with humans reviewing candidates. It also applies when the ground truth is unknown (e.g., recommendation, drug discovery).  

Aligned with classical link prediction metrics such as Hits@1 and Hits@10, we focus on Overlap@1 and Overlap@10.  
We apply this measure both to the top-$K$ predictions (Pred-Overlap@$K$) and to the $K$ nearest neighbors in the embedding space (Space-Overlap@$K$), as detailed below.

\paragraph{Pred-Overlap@$K$.}

Link Prediction–based KG completion can be assumed as populating the KG by adding, for each incomplete query $(h,r,?)$ or $(?,r,t)$, the top-$K$ predicted entities, in line with usual evaluation metrics such as Hits@$K$. Pred-Overlap@$K$ thus measures the extent to which two instances of a model would insert the same triples into the KG.

Let us denote $\mathcal{Q}_\text{test}$ all the test queries that can be derived from test triples, \textit{i.e.}, queries $(h, r, ?)$ and $(?, r, t)$ for each test triple $(h, r, t)$. Let $\nu(I, q, K)$ be the top-$K$ predictions returned by instance $I$ for query $q$.  
For two instances $I_1$ and $I_2$, Pred-Overlap@$K$ is defined as:
\begin{equation}
\text{Pred-Overlap@}K(I_1, I_2)
= \frac{1}{|\mathcal{Q}_\text{test}|}
\sum_{q \in \mathcal{Q}_\text{test}}
\frac{\left|\nu(I_1,q,K) \cap \nu(I_2,q,K)\right|}{K}
\end{equation}
A value close to $1$ indicates that $I_1$ and $I_2$ produce almost the same predictions.
To illustrate, in Figure~\ref{fig:Intro}, runs~1 and~2 share two of their top three predicted entities for each query and thus have a Pred-Overlap@3 of $0.67$. 
Runs~2 and~3 share only one entity for the first query and none for the second, resulting in a Pred-Overlap@3 of only $0.17$.

\paragraph{Space-Overlap@$K$.}

We also seek to measure the agreement between embedding spaces $E_1$ and $E_2$ learned by two instances $I_1$ and $I_2$.  
Embedding spaces are not directly comparable, since they may differ by various transformations, \textit{e.g.}, translation, rotation or scaling.  
Rather than aligning embeddings, we assess stability through neighborhood preservation.  

For each entity $e_i \in \mathcal{E}$, let $\mathcal{N}_K^{E_j}(e_i)$ denote its $K$ nearest neighbors in the embedding space $E_j$, computed using the $\ell_2$ distance\footnote{Cosine distance was also tested, yielding similar results; see our GitHub repository\textsuperscript{\ref{footnote:github}}}. 
Space-Overlap@$K$ reflects the degree to which local neighbors are shared across instances:
\begin{equation}
\text{Space-Overlap@}K(I_1, I_2)
= \frac{1}{|\mathcal{E}|}
\sum_{e_i \in \mathcal{E}}
\frac{\left|\mathcal{N}_K^{E_1}(e_i) \cap \mathcal{N}_K^{E_2}(e_i)\right|}{K}
\end{equation}

\paragraph{Aggregation over groups.}  
To obtain stability scores for a group $G^{\boldsymbol{C}}_\mathcal{X}$, we compute the above metrics across all pairs $(I_1, I_2) \in G^{\boldsymbol{C}}_\mathcal{X}$, and report the mean and standard deviation. We denote these aggregated versions \(\overline{\text{Pred-Overlap@}K}(G^{\boldsymbol{C}}_\mathcal{X})\) and \(\overline{\text{Space-Overlap@}K}(G^{\boldsymbol{C}}_\mathcal{X})\).

\paragraph{Consistency@K and Homogeneity@K.}
Moving beyond pairwise aggregation, we introduce metrics that quantify stability across all runs jointly for one query $q$.
First, we define \textit{Consistency@$K$} as a direct generalization of the overlap: 
we evaluate for one query $q$ the proportion of entities that appear simultaneously in the top-$K$ lists of \emph{all} instances. Second, we define \textit{Homogeneity@$K$} as a generalization of the Jaccard index.
\[
\text{Consistency@}K(q)
= \frac{\left|\displaystyle\bigcap_{i=1}^{n} \nu(I_i,q,K)\right|}{K} \text{\quad Homogeneity@}K(q)
= \frac{\left|\displaystyle\bigcap_{i=1}^{n} \nu(I_i,q,K)\right|}
       {\left|\displaystyle\bigcup_{i=1}^{n} \nu(I_i,q,K)\right|}
\]

\sloppy While Consistency@$K$ quantifies the agreement on top-$K$ predictions, Homogeneity@$K$ additionally penalizes dispersion across instances.

\paragraph{Aggregation over queries.}  
To aggregate consistency and homogeneity metrics on all queries $q \in \mathcal{Q}$, we compute the mean and standard deviation of the above metrics. We denote these aggregated versions \(\overwave{\text{Consistency@}K}\) and \(\overwave{\text{Homogeneity@}K}\).

\subsubsection{Other Metrics.}
We also considered alternative similarity measures.  
A natural candidate is Jaccard similarity.  
However, under the assumption that $|A|=|B|=K$, Jaccard and Overlap are bijectively related, and thus capture the same phenomenon, but Overlap is more interpretable.  
We also considered Rank-Biased Overlap (RBO), which operates on ranked lists and emphasizes top-ranked agreement, as well as Ambiguity and Discrepancy~\cite{zhu2024multiplicty}, which quantify agreement on ranking the gold truth within the top-K predictions.  
Their description and the obtained results are provided in Appendix~\ref{App:metrics}.
Crucially, all these measures lead to the same conclusion: KGEMs exhibit substantial instability across runs.
\section{Experimentation}
\label{sec:Results}

\subsection{Experimental setting}

\paragraph{Models.}
\label{sec:models}
We select representative KGEMs from different families:  
\textbf{TransE}~\cite{TransE} and \textbf{RotatE}~\cite{RotatE} (geometric-based), \textbf{DistMult}~\cite{DistMult} and \textbf{ComplEx}~\cite{ComplEx} (tensor factorization method), \textbf{ConvE}~\cite{ConvE} (convolutional neural networks), \textbf{RGCN}~\cite{RGCN} (graph neural networks), and a \textbf{Transformer} (transformer) without neighborhood context (Section 2.1 of~\cite{Hitter}).

\paragraph{Datasets.}
\label{sec:dataset}
We evaluate our models on standard KG datasets from the literature. Specifically, we consider \textbf{WN18RR}~\cite{ConvE}, a lexical subset of WordNet characterized by its hierarchical structure; \textbf{FB15k-237}~\cite{FB15k-237} and \textbf{Codex-S}~\cite{codex}, which are subsets of Freebase and Wikidata, respectively, and contain real-world facts. 
Table~\ref{table:datasets-stat} provides their statistics.
For completeness, we additionally report in our GitHub repository\textsuperscript{\ref{footnote:github}} results on the smaller \textbf{Kinship} and \textbf{Nations} datasets, commonly used in prior work and describing familial and political relations, respectively.

\begin{table}
	\centering
    \caption{Dataset statistics.}
    \small
	\begin{tabular}{lrrrrr}
		\toprule
		\multicolumn{1}{c}{Dataset}   &  \multicolumn{1}{c}{\# Entities}  & \multicolumn{1}{c}{\# Relations}  & \multicolumn{1}{c}{\# Edges} & \multicolumn{1}{c}{Mean in-degree} \\ 
		\midrule
		WN18RR    & 40,943 & 11  & 93,003   & 2.72  \\ 
		FB15k-237 & 14,541 & 237 & 310,116  & 20.34 \\ 
        CoDEx-S   & 2,034  & 42  & 36,543   & 32.53 \\
		\bottomrule
	\end{tabular}
	\label{table:datasets-stat}
\end{table}

\sloppy In our experiments, the set $S$ of seed values has been fixed to $\{42, 283, 358, 698, 887\}$, and the hardware $\hardware$ has been fixed to a GPU \textit{Nvidia GeForce RTX 2080 Ti (11 GiB)}, except for the specific hardware-variance analysis conducted in Section~\ref{sec:RQ2}.

PyKEEN and LibKGE do not explicitly separate the different sources of randomness, which was necessary for our controlled experiments. Therefore, we implemented a minimal KGE library that provides fine-grained and independent control over the four sources of randomness. It is built on PyTorch, and the RGCN layers are imported from PyTorch Geometric.
Our implementation and the used datasets are publicly accessible\footnote{\label{footnote:github}\url{https://github.com/Wimmics/LinkPerdition}}.

\paragraph{Hyperparameter Search.}
\label{sec:tuning}
The purpose of this tuning step is not to identify the optimal model configuration, a task that would require exploring a substantially larger search space~\cite{OldDog}, but rather to provide a meaningful spectrum of models with different LP performance levels to answer RQ3.

We fixed the seed configuration to $\mathfrak{S} = \{ \mathfrak{S}_\mathcal{N} = s_1,\; \mathfrak{S}_\mathcal{O} = s_1,\; \mathfrak{S}_\mathcal{I} = s_1,\; \mathfrak{S}_\mathcal{D} = s_1 \}$.
We fixed the dropout rates for entities and relations to $0.2$, the batch size to $256$, and the number of negative samples to $500$. 
We used Xavier normal for initialization of weight, and we employ the cross-entropy loss, as \cite{OldDog} shows that this choice consistently yields reliable results.
More details are provided on GitHub\textsuperscript{\ref{footnote:github}}.

Subsequently, we conducted a compact hyperparameter search over embedding dimensions $\{128, 256, 512\}$ and learning rates $\{10^{-6}, 10^{-5}, 10^{-4}, 10^{-3}, 10^{-2}, 10^{-1}\}$.
The resulting configurations were ranked by increasing MRR and those failing to meet a minimum performance threshold ($\text{MRR} < 0.05$) were discarded. We then selected three representative configurations: the best-performing model $\boldsymbol{C_B}$ (highest MRR), the worst-performing model $\boldsymbol{C_W}$ (lowest MRR), and the median-performing model $\boldsymbol{C_M}$ located at the median position of the list.

Note that the best configurations we identified obtain, almost unanimously, performances that lie within a range comparable to state-of-the-art results (see our GitHub repository for the detailed comparison\textsuperscript{\ref{footnote:github}}).

\subsection{Result Analysis}
\subsubsection{RQ1: Stability w.r.t. Randomness.}
\label{sec:RQ1}
If one only considers aggregate scores such as MRR, KGE models appear to be remarkably stable.
Table~\ref{tab:mrr_all} reports the mean MRR and standard deviation across five runs in $G^{\boldsymbol{C_B}}_\mathcal{A}$ (i.e., the best model with all seed components varying jointly). 
Deviations appear consistently small, suggesting that models trained under the same hyperparameter configuration but with varying seeds still achieve comparable overall predictive performance measured by the aggregated ranks of the gold-truth entities.

\begin{table}
\centering
\caption[]{Mean and standard deviation of MRR over the five runs in $G^{\boldsymbol{C_B}}_\mathcal{A}$, \textit{i.e.}, the best model configuration with all seeds jointly varying over $S$\footnotemark.}
\small
\setlength{\tabcolsep}{3pt}
\resizebox{\textwidth}{!}{\begin{tabular}{lccccccc}
\toprule
Dataset \ & TransE \ & ConvE \ & DistMult \ & Transformer \ & RGCN \ & ComplEx \ & RotatE \\ 
\midrule
WN18RR \ & $0.194 \pm 0.002$ \ & $0.410 \pm 0.003$ \ & $0.422 \pm 0.001$ \ & $0.269 \pm 0.010$ \ & $0.415 \pm 0.001$ \ & $0.435 \pm 0.001$ \ & $0.326 \pm 0.039$ \\ 
FB15k-237 \ & $0.315 \pm 0.001$ \ & $0.324 \pm 0.001$ \ & $0.312 \pm 0.002$ \ & $0.295 \pm 0.001$ \ & N/A \ & $0.315 \pm 0.001$ \ & $0.288 \pm 0.001$ \\ 
codex-s \ & $0.348 \pm 0.001$ \ & $0.434 \pm 0.003$ \ & $0.413 \pm 0.003$ \ & $0.360 \pm 0.006$ \ & $0.350 \pm 0.017$ \ & $0.396 \pm 0.002$ \ & $0.359 \pm 0.003$ \\ 
\bottomrule
\end{tabular}}

\label{tab:mrr_all}
\end{table}

\footnotetext{\label{footnote:rgcn}Results for RGCN on FB15k-237 are omitted; under our unified pipeline, this configuration exceeds the memory budget of the GPU used.}

\begin{figure}[h]
    \centering
    \begin{subfigure}{0.48\linewidth}
        \centering
        \includegraphics[width=\linewidth]{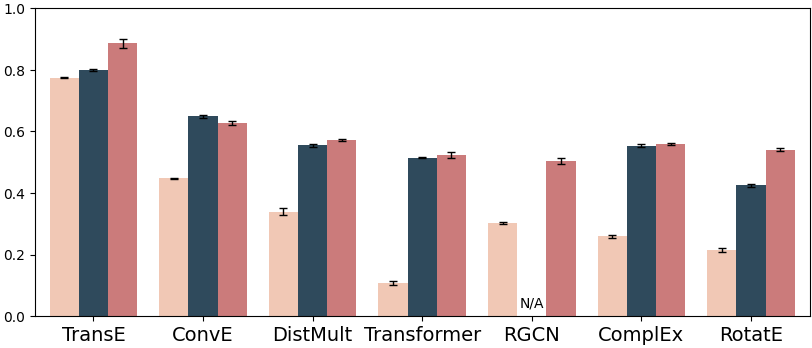}
        \caption{$\overline{\text{Pred-Overlap}@10}$}
        \label{fig:rq1_pred_overlap10}
    \end{subfigure}
    \hfill
    \begin{subfigure}{0.48\linewidth}
        \centering
        \includegraphics[width=\linewidth]{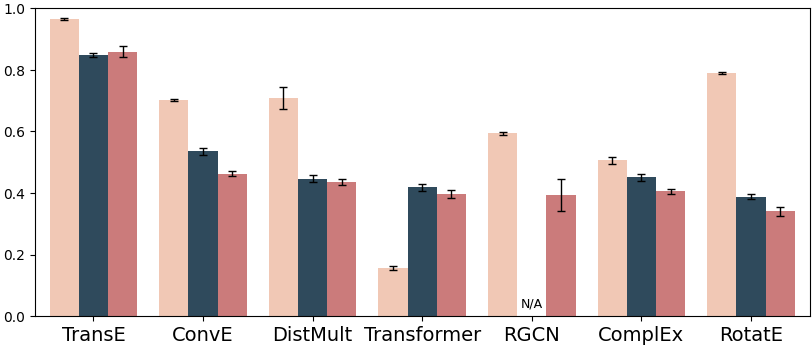}
        \caption{$\overline{\text{Pred-Overlap}@1}$}
        \label{fig:rq1_pred_overlap1}
    \end{subfigure}
    
    \vskip\baselineskip

    \begin{subfigure}{0.48\linewidth}
        \centering
        \includegraphics[width=\linewidth]{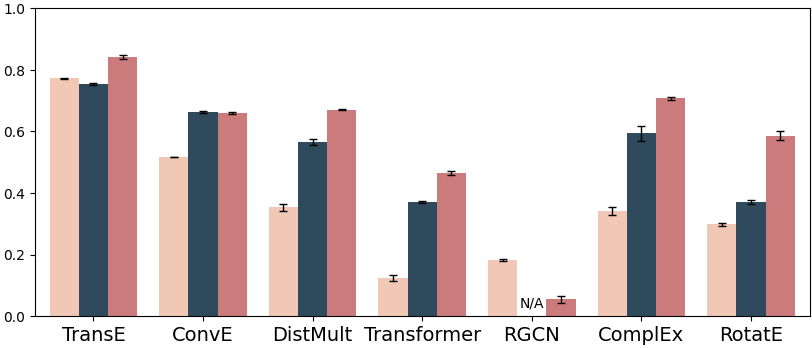}
        \caption{$\overline{\text{Space-Overlap}@10}$}
        \label{fig:rq1_space_overlap10}
    \end{subfigure}

    \caption[]{Stability analysis across all datasets and models\textsuperscript{\ref{footnote:rgcn}}.
    Colors represent the datasets:    
    \raisebox{0pt}{\tikz{\node[draw, rectangle, fill={rgb,255:red,251;green,196;blue,171}, minimum width=14pt, minimum height=6pt] {};}}~WN18RR 
    \raisebox{0pt}{\tikz{\node[draw, rectangle, fill={rgb,255:red,40;green,75;blue,99}, minimum width=14pt, minimum height=6pt] {};}}~FB15k-237 
    \raisebox{0pt}{\tikz{\node[draw, rectangle, fill={rgb,255:red,216;green,110;blue,110}, minimum width=14pt, minimum height=6pt] {};}}~CoDEx-S.
    }
   
    \label{fig:rq1_combined}
\end{figure}

However, our results show that small variations in MRR say little about the stability of individual predictions, and that aggregate metrics mask substantial variability both within and across queries. In Figures~\ref{fig:rq1_pred_overlap10} and \ref{fig:rq1_pred_overlap1}, prediction stability is visualized through $\overline{\text{Pred-Overlap}@10}$ and $\overline{\text{Pred-Overlap}@1}$. For example, two RotatE instances trained with different seeds share on average only $4.3$ of the top-$10$ predicted entities on FB15k-237, dropping to $2.3$ on WN18RR. A similar pattern holds for the top-$1$ prediction, where overlaps are close to or below $0.6$ for most (model, dataset) pairs, meaning that even when completing the KG with only the top-ranked entity, roughly $40\%$ of triples differ across instances. 
Note that the small standard deviations suggest that the magnitude of divergence between any two instances is itself remarkably stable across all pairs.

\begin{figure}[h]
    \centering
    \begin{subfigure}{0.24\linewidth}
        \centering
        \includegraphics[width=\linewidth]{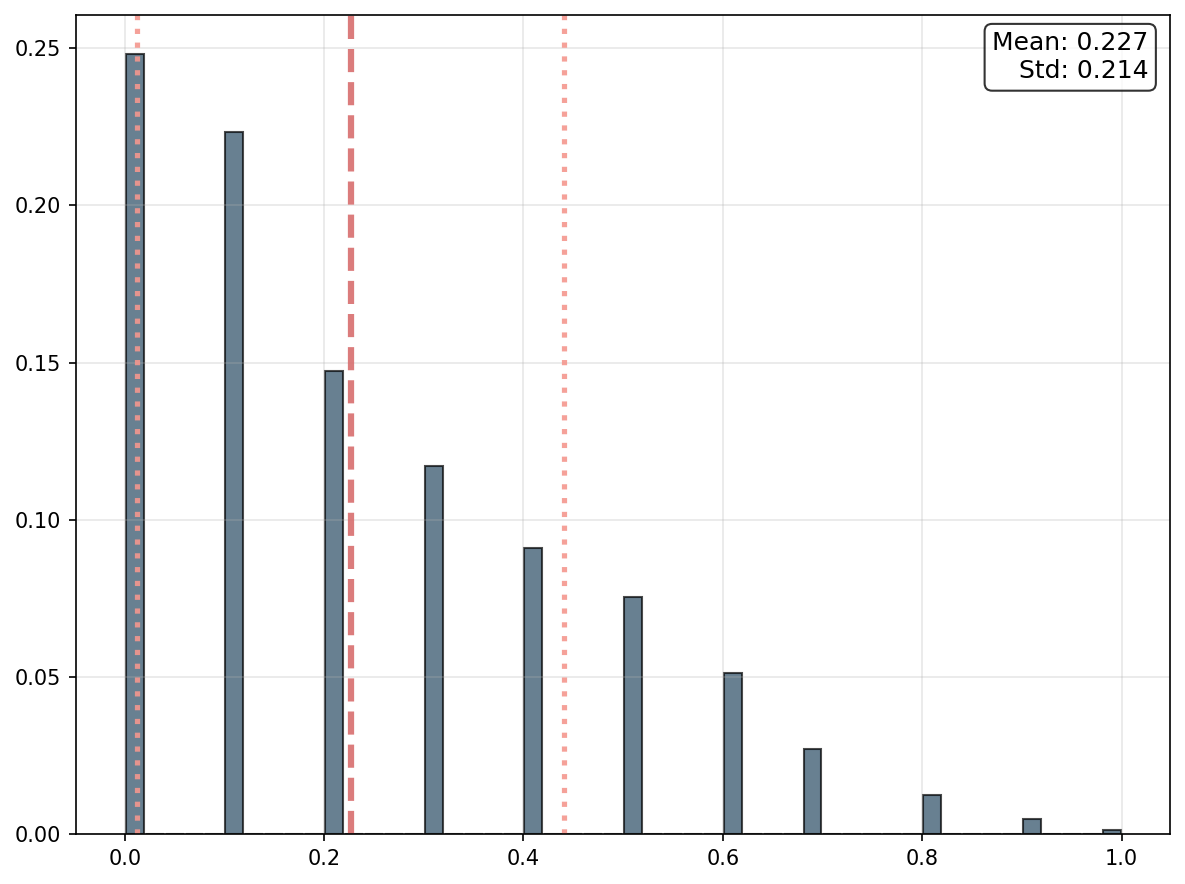}
        \caption{C@10-RotatE}
        \label{fig:ConsistancyRotatE}
    \end{subfigure}
    \hfill
    \begin{subfigure}{0.24\linewidth}
        \centering
        \includegraphics[width=\linewidth]{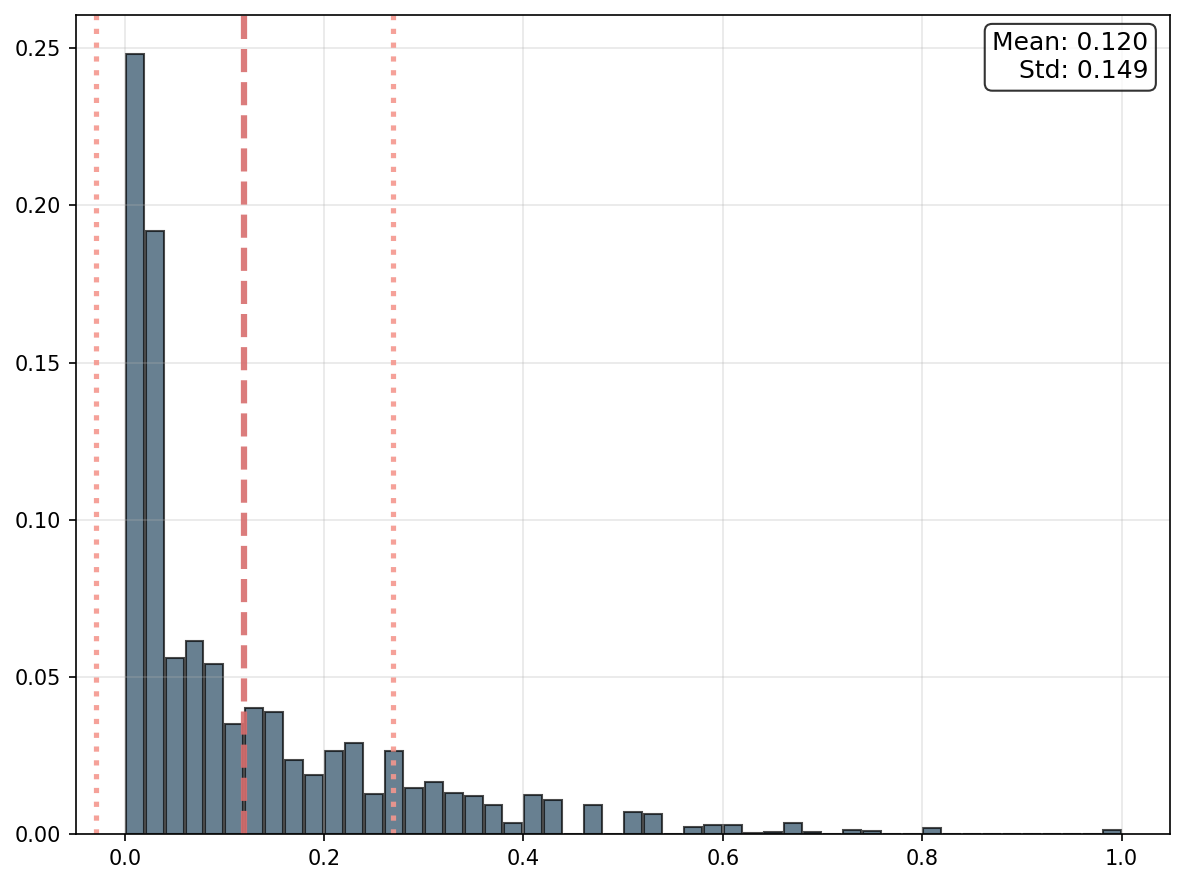}
        \caption{H@10-RotatE}
        \label{fig:HomoRotatE}
    \end{subfigure}
    \hfill
    \begin{subfigure}{0.24\linewidth}
        \centering
        \includegraphics[width=\linewidth]{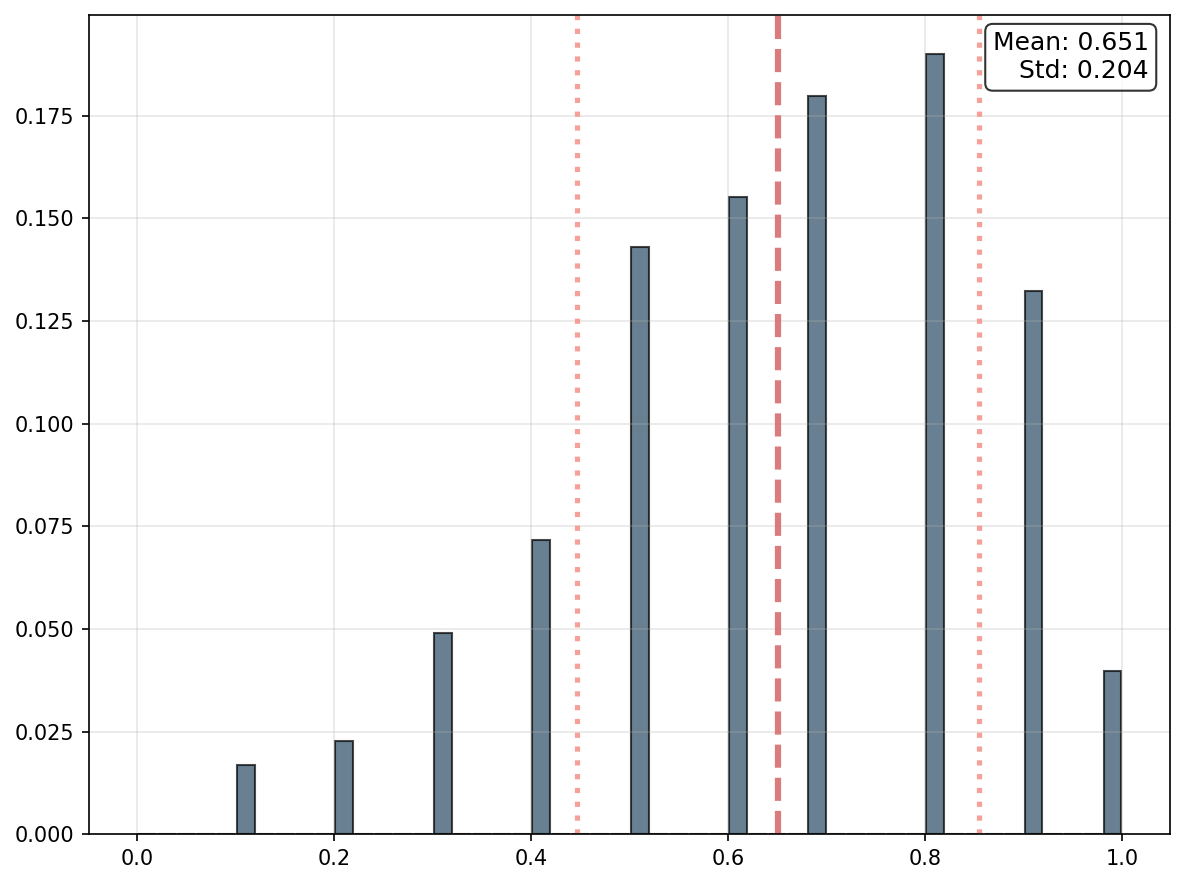}
        \caption{C@10-TransE}
        \label{fig:ConsistancyTransE}
    \end{subfigure}
    \hfill
    \begin{subfigure}{0.24\linewidth}
        \centering
        \includegraphics[width=\linewidth]{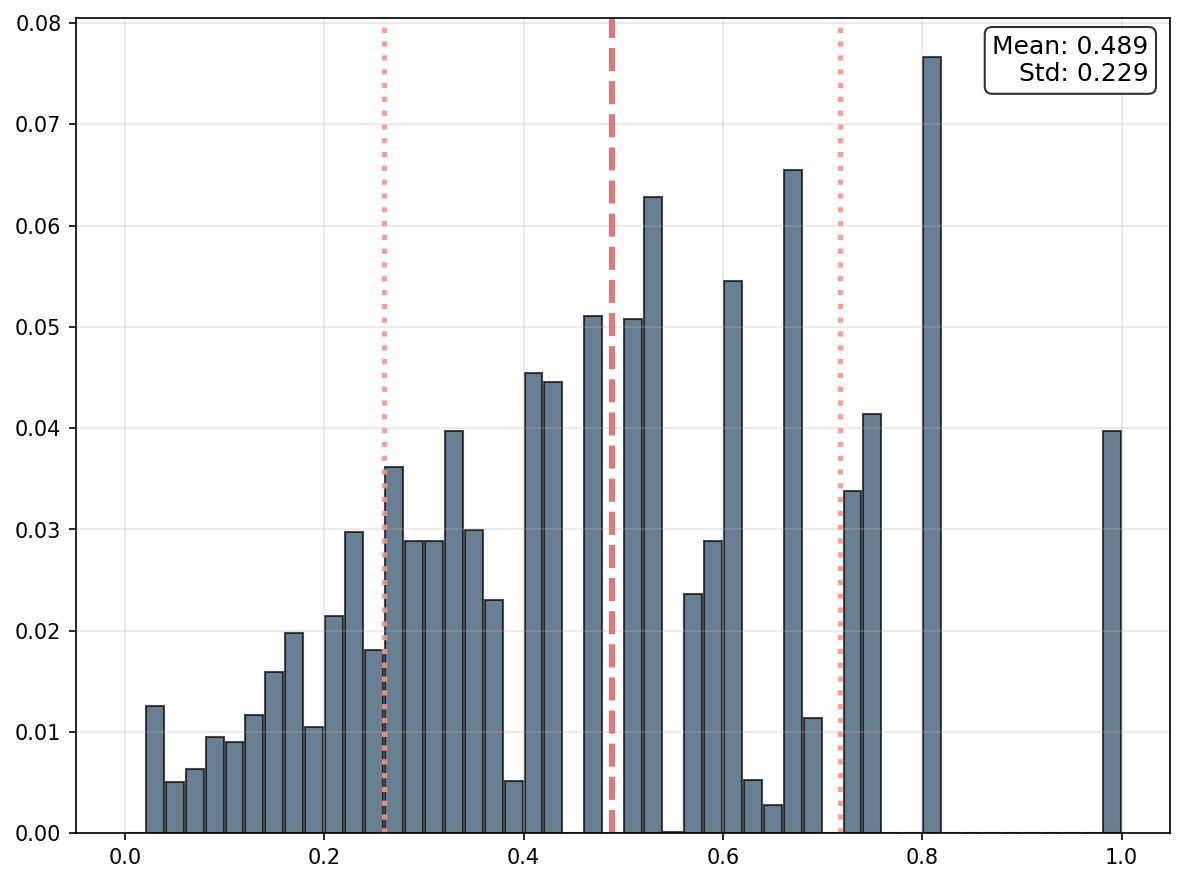}
        \caption{H@10-TransE}
        \label{fig:HomoTransE}
    \end{subfigure}

    \caption{Distribution of Consistency@10 and Homogeneity@10 across queries on FB15k-237 for RotatE and TransE.}
   
    \label{fig:ConstAndHom}
\end{figure}

To further examine variability at the query level, Figure~\ref{fig:ConstAndHom} reports the distributions of Consistency@10 and Homogeneity@10 for TransE and RotatE over WN18RR. Other results are available on GitHub\textsuperscript{\ref{footnote:github}}. For RotatE on FB15k-237, Figure~\ref{fig:ConsistancyRotatE} shows that nearly $25\%$ of queries have a Consistency@10 of $0$, meaning that no entity is common to the top-10 of all five runs, and only $54$ queries (out of $40\,932$) share an identical top-10. Homogeneity@10 provides an even stricter perspective: as shown in Figure~\ref{fig:HomoRotatE}, scores are heavily skewed toward low values, with only $3\%$ of queries exceeding $0.5$, indicating substantial dispersion among non-shared entities.

This behavior is characteristic of most models and datasets. TransE, however, stands out as notably more stable: according to Figure~\ref{fig:rq1_combined}, its Overlap@1 consistently falls between $0.8$ and $1$, and its Overlap@10 comprises on average nearly eight shared entities across runs. This trend is further confirmed by Figures~\ref{fig:ConsistancyTransE} and~\ref{fig:HomoTransE}, although $16\%$ of queries still exhibit a Consistency@10 below or equal to $0.5$.

This instability is not limited to predicted rankings but also manifests in the organization of the embedding space itself, visible through the complementary perspective of $\overline{\text{Space-Overlap}@10}$ in Figure~\ref{fig:rq1_space_overlap10}. 
For instance, an entity shares on average only $5.6$ common neighbors among its 10 nearest ones when comparing two independently trained DistMult instances on FB15k-237.

Taken together, variability in both embedding neighborhoods and predicted rankings indicates that models learn different latent structures and therefore encode different knowledge. Yet none of this variability is reflected in standard LP metrics such as MRR, which depend only on the predicted rank of gold entities. As a result, models with similar MRR can produce substantially divergent completed KGs.
This observation underscores the need for complementary evaluation metrics to more accurately assess KGEM behaviour, a requirement that is particularly important in applications where KGs support decision-making.

\subsubsection{RQ2: Impact of Individual Sources of Randomness.}
\label{sec:RQ2}

\begin{figure}[t]
    \centering
    \begin{subfigure}{0.49\linewidth}
        \centering
        \includegraphics[width=\linewidth]{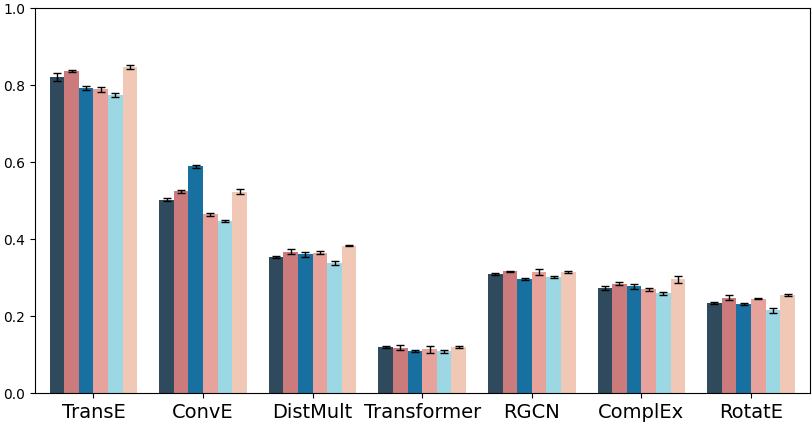}
        \caption{$\overline{\text{Pred-Overlap}@10}$}
        \label{fig:rq2_variant_pred_overlapat10_WN18RR}
    \end{subfigure}
    \begin{subfigure}{0.49\linewidth}
        \centering
        \includegraphics[width=\linewidth]{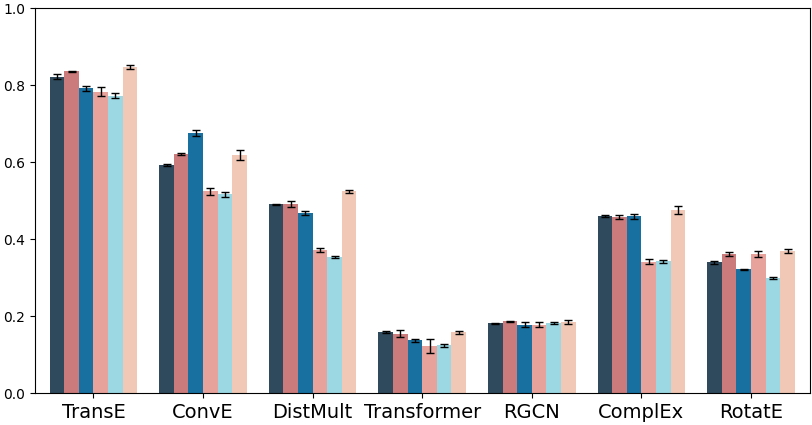}
        \caption{$\overline{\text{Space-Overlap}@10}$}
        \label{fig:rq2_variant_space_overlap10_mean_WN18RR}
    \end{subfigure}

    \caption[]{Impact of individual sources of randomness on WN18RR. Each source of randomness (including hardware) is varied independently while keeping all others fixed. 
    Colors indicate the corresponding comparison group:
    \raisebox{0pt}{\tikz{\node[draw, rectangle, fill={rgb,255:red,40;green,75;blue,99}, minimum width=14pt, minimum height=6pt] {};}}~$G^{\boldsymbol{C_B}}_\mathcal{I}$ (init)
    \raisebox{0pt}{\tikz{\node[draw, rectangle, fill={rgb,255:red,216;green,110;blue,110}, minimum width=14pt, minimum height=6pt] {};}}~$G^{\boldsymbol{C_B}}_\mathcal{D}$ (dropout)
    \raisebox{0pt}{\tikz{\node[draw, rectangle, fill={rgb,255:red,0;green,119;blue,182}, minimum width=14pt, minimum height=6pt] {};}}~$G^{\boldsymbol{C_B}}_\mathcal{N}$ (neg)
    \raisebox{0pt}{\tikz{\node[draw, rectangle, fill={rgb,255:red,244;green,151;blue,142}, minimum width=14pt, minimum height=6pt] {};}}~$G^{\boldsymbol{C_B}}_\mathcal{O}$ (order)
    \raisebox{0pt}{\tikz{\node[draw, rectangle, fill={rgb,255:red,144;green,224;blue,239}, minimum width=14pt, minimum height=6pt] {};}}~$G^{\boldsymbol{C_B}}_\mathcal{A}$ (all)
    \raisebox{0pt}{\tikz{\node[draw, rectangle, fill={rgb,255:red,251;green,196;blue,171}, minimum width=14pt, minimum height=6pt] {};}}~$G^{\boldsymbol{C_B},\mathfrak{S}}_\hardware$ (hardware).
    The category ``all'' includes all sources of randomness but hardware.}
    \label{fig:rq2_variant_WN18RR}
\end{figure}

We now examine the impact of each source of randomness in isolation. 
In particular, we compare stability across $G^{\boldsymbol{C_B}}_\mathcal{N}$, $G^{\boldsymbol{C_B}}_\mathcal{I}$, $G^{\boldsymbol{C_B}}_\mathcal{O}$, 
$G^{\boldsymbol{C_B}}_\mathcal{D}$ (varying random seeds independently for each source of randomness on a fixed hardware), and $G^{\boldsymbol{C_B}}_\mathcal{A}$ (varying all seeds simultaneously on a fixed hardware). 
Figure~\ref{fig:rq2_variant_pred_overlapat10_WN18RR} and Figure~\ref{fig:rq2_variant_space_overlap10_mean_WN18RR} report, for WN18RR, the instability associated with each group, respectively at the prediction level and in the embedding space.
Instability measures of the other considered datasets are available in Figure~\ref{fig:app_rq2_metrics} of Appendix~\ref{App:RQ2}.
Our results indicate that, for each model, each source independently is sufficient to induce instability at  both the embedding and prediction levels. 
This highlights the inherent and persistent instability induced by randomness factors in model training.  
Across all datasets and metrics, no particular seed emerges as a dominant factor of instability. 
The observed variations in instability appear to be more strongly tied to the model architecture than to the specific source of randomness.
Note that the comparable impact of each source of randomness validates the design choice of frameworks such as PyKeen and LibKGE to expose only a single seed controlling all stochastic factors.

Furthermore, we define an additional group $G^{\boldsymbol{C_B},\mathfrak{S}}_\hardware$, in which the hyperparameter and seed configurations $\boldsymbol{C_B}$ and $\mathfrak{S}$ are fixed, while the hardware varies.  
Each model in this group has been trained on a different GPU: \textit{Nvidia GeForce RTX 2080 Ti (11 GiB)}, \textit{Nvidia GeForce GTX 1080 Ti (11 GiB)}, \textit{Nvidia Tesla T4 (15 GiB)}, \textit{Nvidia A40 (45 GiB)}, and \textit{Nvidia A100-SXM4-40GB (40 GiB)}. 
Strikingly, this setting exhibits instability of the same magnitude as seed variation (in beige in~\ref{fig:rq2_variant_space_overlap10_mean_WN18RR}), highlighting the challenge of ensuring reproducibility in both research and production environments, particularly in cloud computing scenarios where hardware choice may be beyond the user’s control.

\subsubsection{RQ3: Relationship Between Stability and Link Prediction Performance.}
\label{sec:RQ3}

We now investigate the model-wise relationship between the predictive performance (MRR) of a configuration and its stability. We aim to identify, for each model, which of $G^{\boldsymbol{C_B}}_\mathcal{A}$, $G^{\boldsymbol{C_M}}_\mathcal{A}$, and $G^{\boldsymbol{C_W}}_\mathcal{A}$, respectively the best, medium, and worst configurations from the hyperparameter search, exhibits the greatest instability.

\begin{figure}[t]
    \centering

    \begin{subfigure}{0.41\linewidth}
        \centering
        \includegraphics[height=6cm]{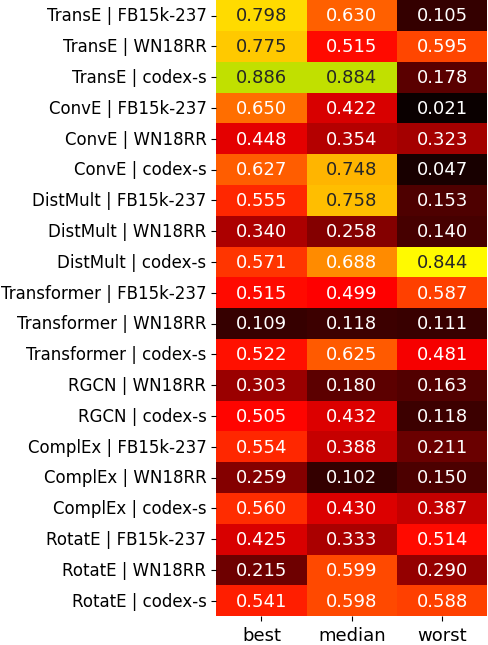}
        \caption{$\overline{\text{Pred-Overlap@10}}$}
        \label{fig:rq3_variant_pred_overlapat10}
    \end{subfigure}\hfill
    \begin{subfigure}{0.26\linewidth}
        \centering
        \includegraphics[height=6cm]{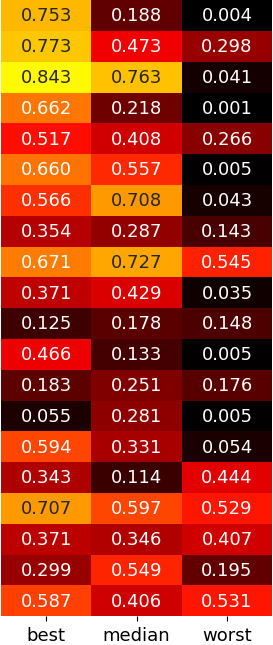}
        \caption{$\overline{\text{Space-Overlap@10}}$}
        \label{fig:rq3_variant_space_overlapat10_mean}
    \end{subfigure}\hfill
    \begin{subfigure}{0.26\linewidth}
        \centering
        \includegraphics[height=6cm]{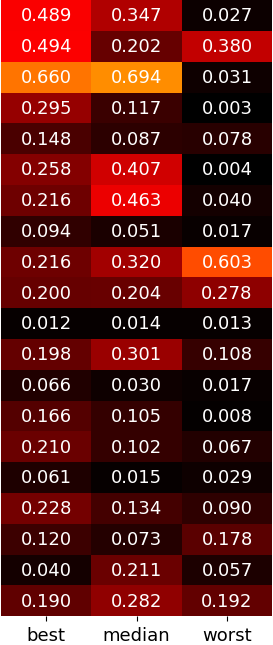}
        \caption{$\overwave{\text{Homogeneity@10}}$}
        \label{fig:rq3_variant_super_pred_jaccardat10_mean}
    \end{subfigure}
    \begin{subfigure}{0.05\linewidth}
        \centering
        \raisebox{1.2cm}{\includegraphics[height=5cm]{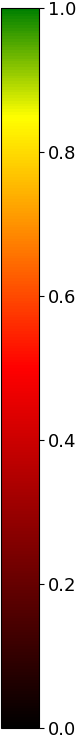}}
    \end{subfigure}
    \caption{Impact of model performance on stability.}
    \label{fig:rq3_combined}
\end{figure}

Our observations (Figures~\ref{fig:rq3_variant_pred_overlapat10} and~\ref{fig:rq3_variant_space_overlapat10_mean}) show that poorly performing model configurations ($G^{\boldsymbol{C_W}}_\mathcal{A}$) tend to be less stable, which aligns with intuition. 
Nevertheless, counterintuitively, we observe no clear relationship between performance and stability between high ($G^{\boldsymbol{C_B}}_\mathcal{A}$) and median ($G^{\boldsymbol{C_M}}_\mathcal{A}$) performing model configurations. 
At both the embedding and prediction levels, high-performing model configurations do not consistently exhibit greater stability than median-performing ones. 
Figure~\ref{fig:rq3_combined} thus suggests that stability is largely independent of configuration \textit{performance}, except for poorly trained instances, and is more strongly influenced by model architecture.

\subsection{Complementary Study: Voting as an Instability Remediation?}

Voting was proposed by Zhu et al.~\cite{zhu2024multiplicty} to alleviate instability, motivating us to evaluate its effect on our proposed stability metrics. In particular, we used Borda Vote and Range Vote.

For this experiment, for each model, we reused the model instances trained in the previous sections and randomly partitioned them into five bins of five instances each, independently of their original group $G^{\boldsymbol{C}}_\mathcal{X}$. 
Within each bin, we applied the voting procedure, yielding one aggregated model per bin for each voting method. 
We then computed the stability measures between these aggregated models, for each of the two voting procedures.

\begin{figure}[h]
    \centering

    \begin{subfigure}[t]{0.48\linewidth}
        \centering
        \includegraphics[width=\linewidth]{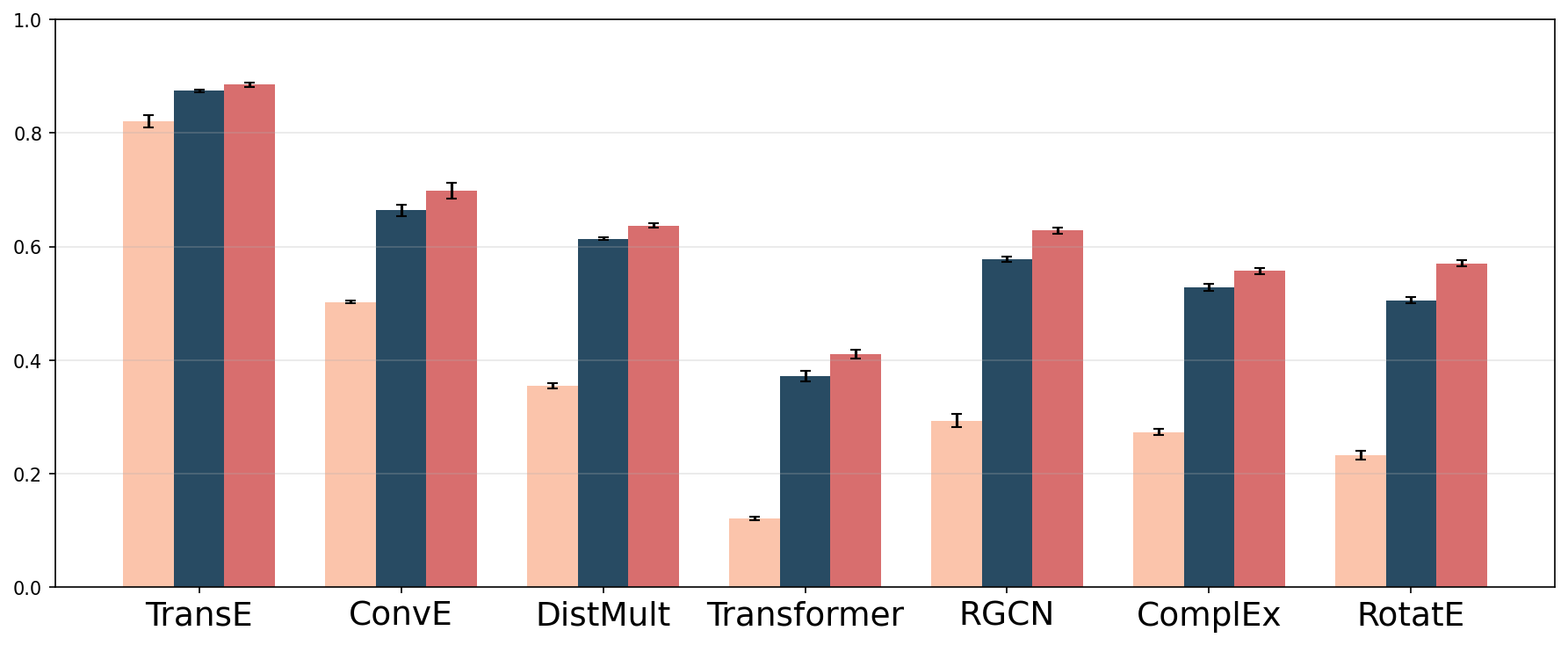}
        \caption{WN18RR}
        \label{fig:voting_wn18rr}
    \end{subfigure}
    \hfill
    \begin{subfigure}[t]{0.48\linewidth}
        \centering
        \includegraphics[width=\linewidth]{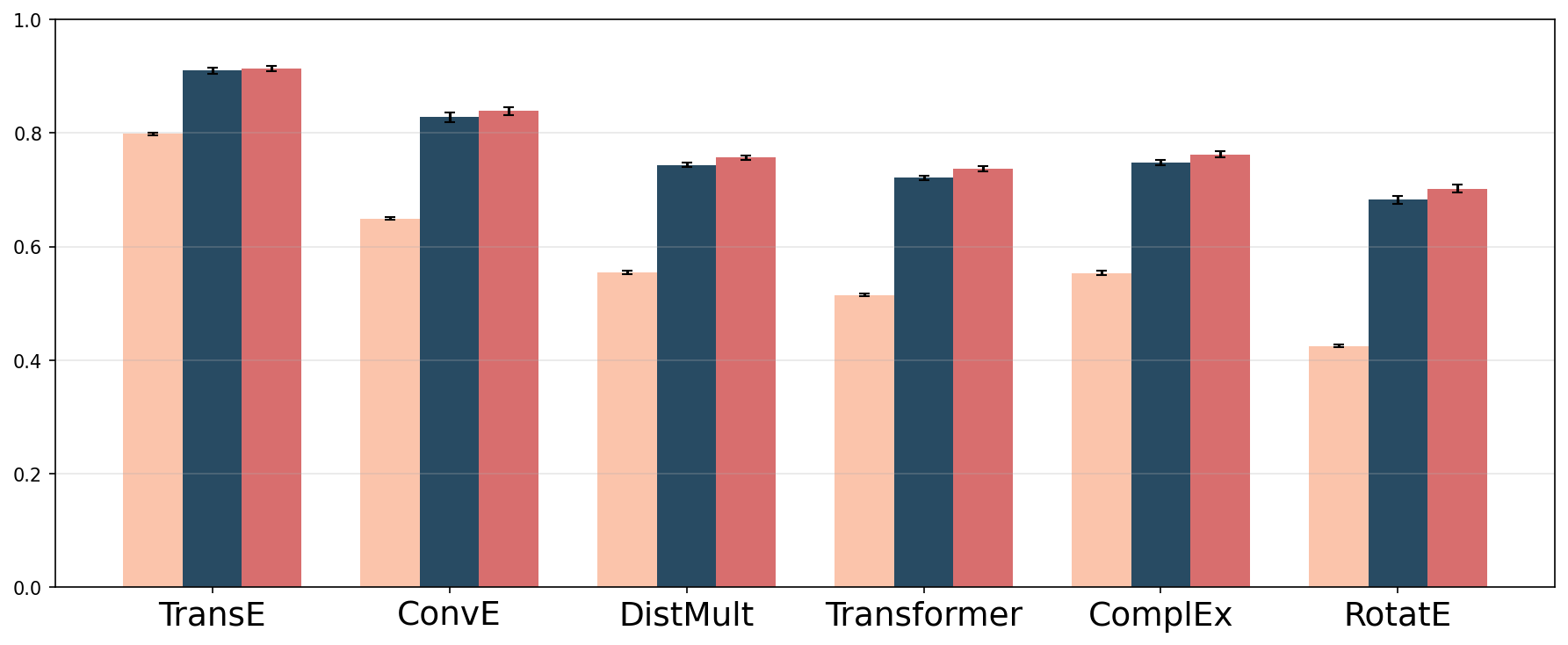}
        \caption{FB15k-237}
        \label{fig:voting_fb15k}
    \end{subfigure}
    \vskip\baselineskip
    \begin{subfigure}[t]{0.48\linewidth}
        \centering
        \includegraphics[width=\linewidth]{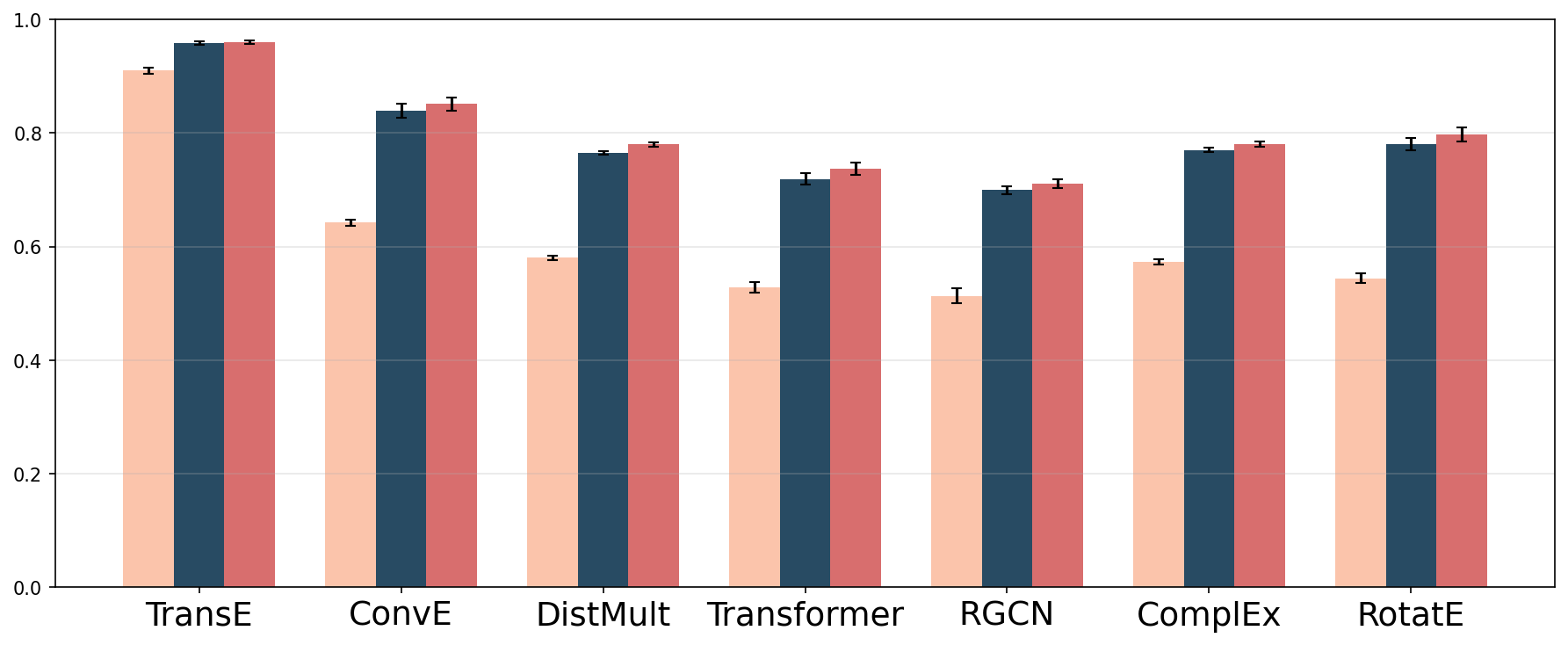}
        \caption{CoDEx-S}
        \label{fig:voting_codexs}
    \end{subfigure}

    \caption[]{Improvement on $\overline{\text{Pred-Overlap}@10}$ brought by voting across all models and datasets.       
    Colors represent the voting method: 
    \raisebox{0pt}{\tikz{\node[draw, rectangle, fill={rgb,255:red,251;green,196;blue,171}, minimum width=14pt, minimum height=6pt] {};}}~No voting baseline ($G^{\boldsymbol{C}}_\mathcal{A}$ instability)
    \raisebox{0pt}{\tikz{\node[draw, rectangle, fill={rgb,255:red,40;green,75;blue,99}, minimum width=14pt, minimum height=6pt] {};}}~Borda 
    \raisebox{0pt}{\tikz{\node[draw, rectangle, fill={rgb,255:red,216;green,110;blue,110}, minimum width=14pt, minimum height=6pt] {};}}~Score. 
    }
   
    \label{fig:voting}
\end{figure}

Figure~\ref{fig:voting} displays the obtained results, as well as $\overline{\text{Pred-Overlap}@10}$ obtained on $G^{\boldsymbol{C}}_\mathcal{A}$ as a baseline instability. 
It demonstrates that voting improves stability across all models. 
However, the instability is reduced but not eliminated, and voting comes with additional computational cost since it requires training multiple instances of the same model. 
Moreover, although voting can be straightforwardly used on predictions, it is not applicable to embedding-space neighborhoods. 
Hence, instability at the embedding space level remains an unaddressed issue.

\section{Discussion and Conclusion}
\label{sec:conclusions}

We conducted a systematic stability analysis of several KGEMs across multiple datasets, focusing on the impact of random seeds (initialization, triple ordering, negative sampling, and dropout) as well as hardware (e.g., GPUs) on KGEMs overall performance, predictions, and embedding space organization. 
Although aggregate metrics such as MRR and Hits@$K$ appear stable, they conceal pronounced instabilities. At the prediction level, models produce divergent top-$K$ candidates for identical incomplete triples. 
At the embedding level, latent neighborhoods are also seed-dependent. Our observations connect with prior work \cite{DoSimilar,DoSemantic}, which cautioned that embeddings do not necessarily preserve semantics or graph proximity. This underscores the risks of treating KGEMs as faithful neural counterparts of KGs and questions the validity of downstream operations on the embedding space such as similarity measurement, clustering, neighborhood-based or space-based reasoning~\cite{MonninRNC22,abs-2304-11949,ZhangPWCZZBC19}. 
Together, these observations raise concerns for KG completion and downstream applications. Seed-dependent variability may undermine the reliability of the systems that rely on completed KGs or KGEMs for decision making.

Our analysis shows that altering any of these factors is sufficient to induce comparable instability, leaving reproducibility highly sensitive to seemingly minor stochastic choices. 
Worrisomely, the impact of GPU types on stability raises additional concerns for reproducibility, particularly in cloud or shared computing environments where users have limited control over the underlying hardware.
We also show that performance and stability are not aligned criteria, highlighting the need to treat them as distinct evaluation dimensions for model selection. This motivates the design and adoption of broader evaluation protocols for link prediction that reflect not only benchmark performance but also the robustness required in practical applications.

For future work, our results indicate that certain models, such as TransE, exhibit relatively higher stability. This suggests a promising avenue for designing new architectures or training procedures (e.g., stability-aware losses) that explicitly enforce stability. To do so, it would be valuable to analyze the impact of hyperparameters (such as learning rate, dropout rate or embedding dimension\gmt{s}) on stability.
Another direction lies in exploring whether other KGEM formalisms like RDF2Vec~\cite{RDF2Vec}, NBFNet~\cite{NBFNET}, or MINERVA~\cite{MINERVA}, rule-mining systems such as AMIE+~\cite{AmiePlus} or AnyBURL~\cite{AnyBURL}, or hybrid approaches like IterE~\cite{ZhangPWCZZBC19}, provide more stable predictions and spaces.

\section*{Acknowledgement}
This work has been supported by the French government, through the 3IA Côte d’Azur Investments in the project managed by the National Research Agency (ANR) with the reference number ANR-23-IACL-0001.    
Experiments presented in this paper were carried out using the Grid'5000 testbed, supported by a scientific interest group hosted by Inria and including CNRS, RENATER and several Universities as well as other organizations (see \url{https://www.grid5000.fr}). 
This publication is based upon work from COST Action CA23147 GOBLIN - Global Network on Large-Scale, Cross-domain and Multilingual Open Knowledge Graphs, supported by COST (European Cooperation in Science and Technology, \url{https://www.cost.eu}).

\section*{Supplemental Material Statement}
To ensure the reproducibility of our findings, all code and datasets (WN18RR, FB15k-237, CoDEx-s, kinship and nations) are provided in our GitHub repository\textsuperscript{\ref{footnote:github}}, allowing to reproduce training, evaluation, and similarity computations. 
Hyperparameter configurations of grid search and selected models are documented in the README. 
We also precisely report the hardware (GPUs) and seeds employed in each experiment in the paper.

\section*{Use of Generative AI}
In the preparation of this paper, ChatGPT was used as a writing assistant for grammar and spelling checks. In addition, Windsurf Editor, powered by Claude 3.5, was used as a coding assistant for implementation details.
All scientific contributions, analyses, and experimental results presented in this paper are the authors' own.

%
%
\bibliographystyle{splncs04}
\bibliography{mybibliography}
%

\newpage

\appendix
\section*{APPENDICES}

\section{Other Metrics}
\label{App:metrics}

\textbf{Jaccard} similarity is a natural candidate. However, under the assumption $|A|=|B|=K$, it is bijectively related to Overlap:
\begin{equation}
\text{Jaccard}(A,B)
= \frac{|A\cap B|}{|A\cup B|}
= \frac{|A\cap B|}{2K - |A\cap B|}
= \frac{\text{Overlap@}K(A,B)}{2 - \text{Overlap@}K(A,B)}.
\end{equation}

\noindent \textbf{Rank-Biased Overlap} (RBO) weights top-ranked elements more heavily:
\begin{equation}
\text{(Space/Pred)-RBO@}K(I_1, I_2)
= \frac{1}{K} \sum_{d=1}^K 
\text{(Space/Pred)-Overlap@}d(I_1,I_2).
\end{equation}

\noindent \textbf{Ambiguity}, adapted from \cite{zhu2024multiplicty}, measures disagreement on whether the gold entity is ranked within the top-$K$:
\begin{equation}
\text{Ambiguity@}K(G^{\boldsymbol{C}}_\mathcal{X}) =
\frac{1}{|\mathcal{Q}_\text{test}|}
\sum_{q \in \mathcal{Q}_\text{test}}
\max_{I_1,I_2 \in G^{\boldsymbol{C}}_\mathcal{X}}
\Delta(I_1, I_2, q, K).
\end{equation}
where $\Delta(I_1, I_2, q, K)=0$ if both runs place the correct entity either inside or outside the top-$K$, and $1$ otherwise.  
Unlike the other metrics, higher Ambiguity indicates more frequent disagreement; we therefore plot $1 - \text{Ambiguity}@K$ for comparability.  

\begin{figure}[h]
    \centering
    \begin{subfigure}{0.48\linewidth}
        \centering
        \includegraphics[width=\linewidth]{figures/rq1_pred_overlapat10.png}
        \caption{$\overline{\text{Pred-Overlap}@10}$}
        \label{fig:rq1_pred_overlap10_appendix}
    \end{subfigure}
    \hfill
    \begin{subfigure}{0.48\linewidth}
        \centering
        \includegraphics[width=\linewidth]{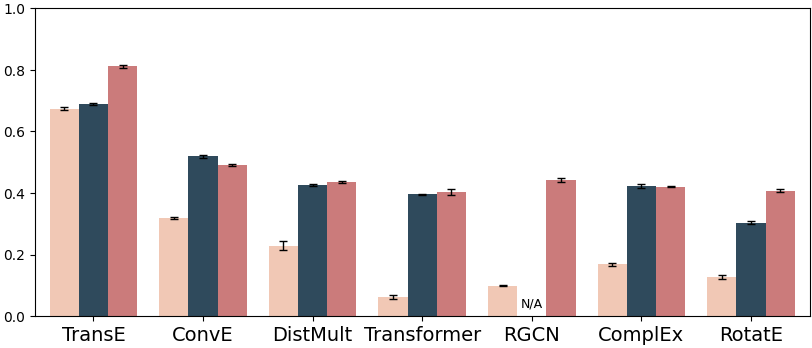}
        \caption{$\overline{\text{Pred-Jaccard@10}}$}
        \label{fig:rq1_pred_jaccard10}
    \end{subfigure}

    \vskip\baselineskip
    \begin{subfigure}{0.48\linewidth}
        \centering
        \includegraphics[width=\linewidth]{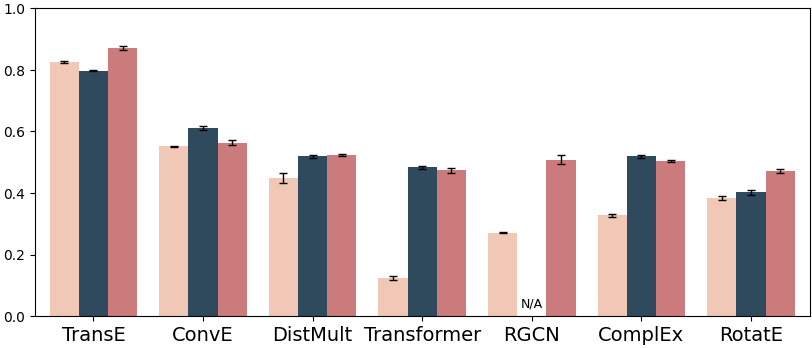}
        \caption{$\overline{\text{Pred-RBO@10}}$}
        \label{fig:rq1_pred_rbo10}
    \end{subfigure}
    \hfill
    \begin{subfigure}{0.48\linewidth}
        \centering
        \includegraphics[width=\linewidth]{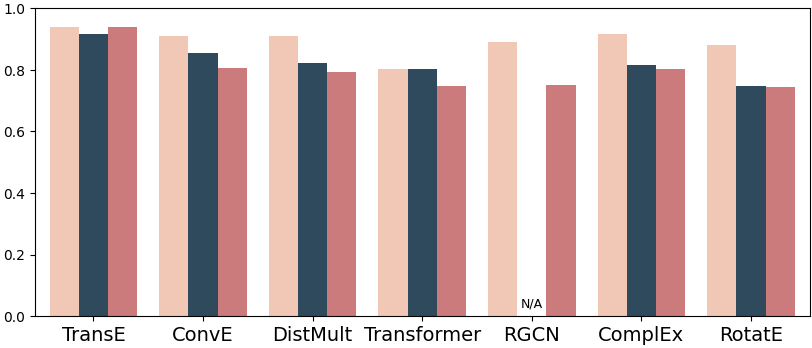}
        \caption{$1 - \text{Ambiguity@}10$}
        \label{fig:rq1_ambiguity10}
    \end{subfigure}
    \caption[]{Comparison of the different metrics across all datasets and models\textsuperscript{\ref{footnote:rgcn}}.
    Colors denote datasets:
    \raisebox{0pt}{\tikz{\node[draw, rectangle, fill={rgb,255:red,251;green,196;blue,171}, minimum width=14pt, minimum height=6pt] {};}}~WN18RR\;
    \raisebox{0pt}{\tikz{\node[draw, rectangle, fill={rgb,255:red,40;green,75;blue,99}, minimum width=14pt, minimum height=6pt] {};}}~FB15k-237\;
    \raisebox{0pt}{\tikz{\node[draw, rectangle, fill={rgb,255:red,216;green,110;blue,110}, minimum width=14pt, minimum height=6pt] {};}}~CoDEx-S.
    }
    \label{fig:pred_met_combined}
\end{figure}

As shown in Figure~\ref{fig:pred_met_combined}, 
$\overline{\text{Pred-Jaccard@10}}$ and $\overline{\text{Pred-RBO@10}}$ follow patterns similar to $\overline{\text{Pred-Overlap@10}}$ across all (model, dataset) pairs.  
$\overline{\text{Pred-RBO@10}}$ yields slightly higher values on WN18RR, suggesting more agreement among top-ranked predictions, though instability remains.
Ambiguity@10 paints a different picture where models mostly agree on placing the gold truth entity either within or out of the top-$K$ predictions. 
That is why, we argue that Ambiguity only captures a restricted view of model instability, motivating the need for our proposed metrics. 

\section{Instability Results for RQ2 on Other Datasets}
\label{App:RQ2}

\begin{figure}[h]
    \centering
    
    \begin{subfigure}{0.48\linewidth}
        \centering
        \includegraphics[width=\linewidth]{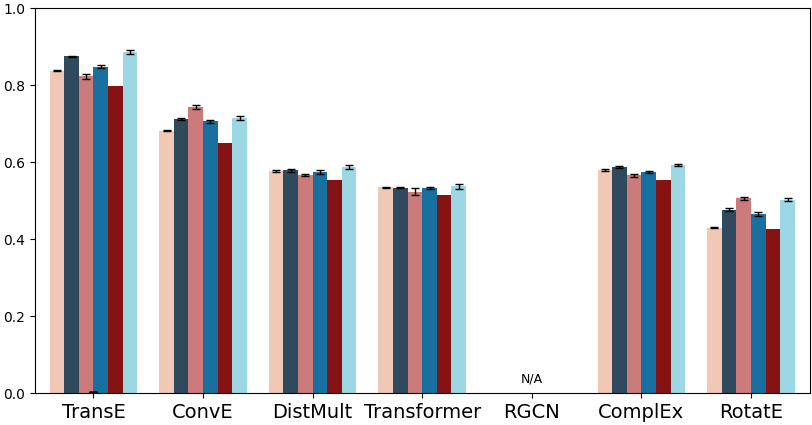}
        \caption{FB15k-237 -- Pred-Overlap@10}
        \label{fig:Apprq2_jaccard10_WN18RR}
    \end{subfigure}
    \hfill
    \begin{subfigure}{0.48\linewidth}
        \centering
        \includegraphics[width=\linewidth]{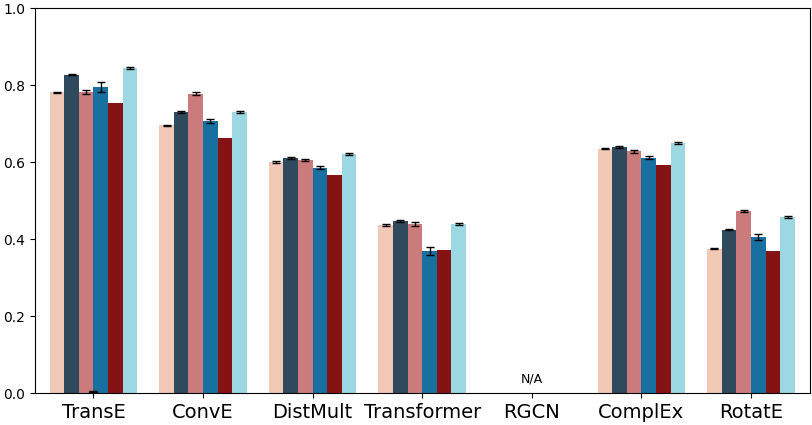}
        \caption{FB15k-237 -- Space-Overlap@10}
        \label{fig:Apprq2_predjaccard1_WN18RR}
    \end{subfigure}

    \begin{subfigure}{0.48\linewidth}
        \centering
        \includegraphics[width=\linewidth]{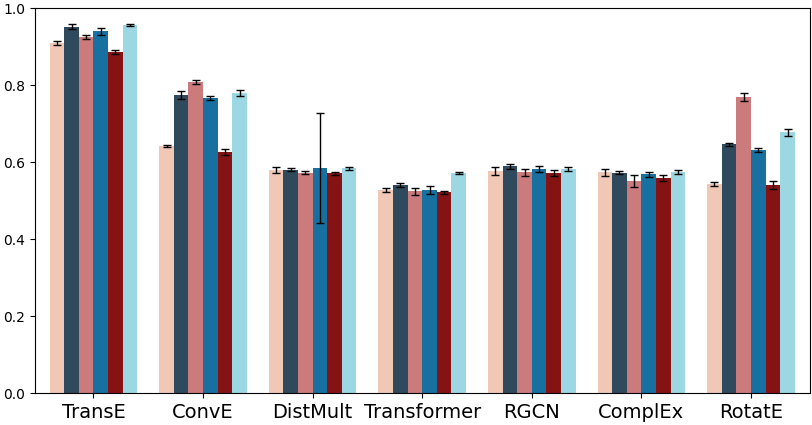}
        \caption{CoDEx-S -- Pred-Overlap@10}
        \label{fig:Apprq2_predoverlap10_codexs}
    \end{subfigure}
    \hfill
    \begin{subfigure}{0.48\linewidth}
        \centering
        \includegraphics[width=\linewidth]{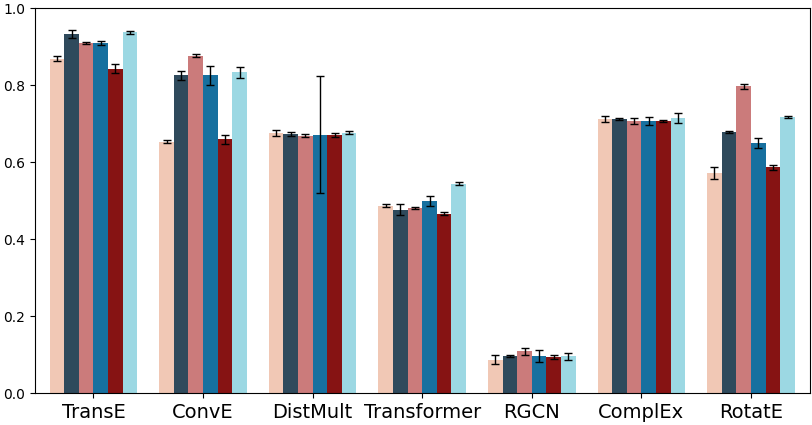}
        \caption{CoDEx-S -- Space-Overlap@10}
        \label{fig:Apprq2_overlap10_codexs}
    \end{subfigure}
    
    \caption[]{Individual impact for each sources of randomness across datasets. Colors indicate the corresponding comparison group: 
    \raisebox{0pt}{\tikz{\node[draw, rectangle, fill={rgb,255:red,251;green,196;blue,171}, minimum width=14pt, minimum height=6pt] {};}}~$G^{\boldsymbol{C_B}}_\mathcal{I}$ (init)
    \raisebox{0pt}{\tikz{\node[draw, rectangle, fill={rgb,255:red,40;green,75;blue,99},     minimum width=14pt, minimum height=6pt] {};}}~$G^{\boldsymbol{C_B}}_\mathcal{D}$ (dropout)
    \raisebox{0pt}{\tikz{\node[draw, rectangle, fill={rgb,255:red,216;green,110;blue,110}, minimum width=14pt, minimum height=6pt] {};}}~$G^{\boldsymbol{C_B}}_\mathcal{N}$ (neg)
    \raisebox{0pt}{\tikz{\node[draw, rectangle, fill={rgb,255:red,0;green,119;blue,182},   minimum width=14pt, minimum height=6pt] {};}}~$G^{\boldsymbol{C_B}}_\mathcal{O}$ (order)
    \raisebox{0pt}{\tikz{\node[draw, rectangle, fill={rgb,255:red,153;green,0;blue,0},     minimum width=14pt, minimum height=6pt] {};}}~$G^{\boldsymbol{C_B}}_\mathcal{A}$ (all)
    \raisebox{0pt}{\tikz{\node[draw, rectangle, fill={rgb,255:red,144;green,224;blue,239}, minimum width=14pt, minimum height=6pt] {};}}~ $G^{\boldsymbol{C_B},\mathfrak{S}}_\hardware$ (hardware).
    The category ``all'' includes all sources of randomness but hardware.
    }
    \label{fig:app_rq2_metrics}
\end{figure}

Figure~\ref{fig:app_rq2_metrics} confirms our analysis in Section~\ref{sec:RQ2} using the other datasets:
 \begin{enumerate}
    \item Each source is independently sufficient to induce instability at both the embedding and prediction levels;
    \item There is an inherent and persistent instability induced by randomness factors in model training;
    \item The observed variations in instability appear to be more strongly tied to the model architecture than to the specific source of randomness.
 \end{enumerate}%


\end{document}